*Original Article*

# Autotelic Reinforcement Learning: Exploring Intrinsic Motivations for Skill Acquisition in Open-Ended Environments

Prakhar Srivastava[1], Jasmeet Singh[2]

[1,2]*University of Illinois at Chicago, Illinois, USA.*

[1]*Corresponding Author : prakharsrivsatava002@gmail.com*

***Abstract -*** *This paper presents a comprehensive overview of autotelic Reinforcement Learning (RL), emphasizing the role of intrinsic motivations in the open-ended formation of skill repertoires. We delineate the distinctions between knowledge-based and competence-based intrinsic motivations, illustrating how these concepts inform the development of autonomous agents capable of generating and pursuing self-defined goals. The typology of Intrinsically Motivated Goal Exploration Processes (IMGEPs) is explored, with a focus on the implications for multi-goal RL and developmental robotics. The autotelic learning problem is framed within a reward-free Markov Decision Process (MDP), WHERE agents must autonomously represent, generate, and master their own goals. We address the unique challenges in evaluating such agents, proposing various metrics for measuring exploration, generalization, and robustness in complex environments. This work aims to advance the understanding of autotelic RL agents and their potential for enhancing skill acquisition in a diverse and dynamic setting.*

***Keywords -*** *Skill Acquisition,  Reinforcement learning, Social autonomous agents, Open-Ended environments, Social Learners.*

## 1. Introduction

Artificial Intelligence (AI) aims to create autonomous agents that can operate across diverse environments and complete a wide range of tasks. Researchers pursue different approaches, each focusing on specific drivers of learning. In Reinforcement Learning (RL) [1], agents learn by exploring their environment and using their experience to solve tasks. Imitation Learning (IL) [2] involves agents learning from expert demonstrations, while Multi-Agent Reinforcement Learning (MARL) [3] emphasizes cooperation among agents to solve collaborative tasks. Recent advancements in RL have demonstrated success in varied domains, such as playing Atari games [4], mastering chess and Go [5], and controlling stratospheric balloons [6]. IL, combined with transformers [7], has enabled generalist agents to be trained on diverse datasets and to perform in-context reinforcement learning via algorithm distillation. However, these algorithms remain sample-inefficient and struggle with generalization, creativity, and tackling novel tasks, largely because they rely on isolated learning signals. This research explores sociocultural interactions as a new avenue for AI learning inspired by human development. By immersing artificial agents in social contexts, we investigate how sociocultural dynamics impact learning. The first part of this study examines the formation of cultural conventions among agents, while the second part introduces a framework called Vygotskian Autotelic Artificial Intelligence, which leverages sociocultural interactions to enhance open-ended skill acquisition.

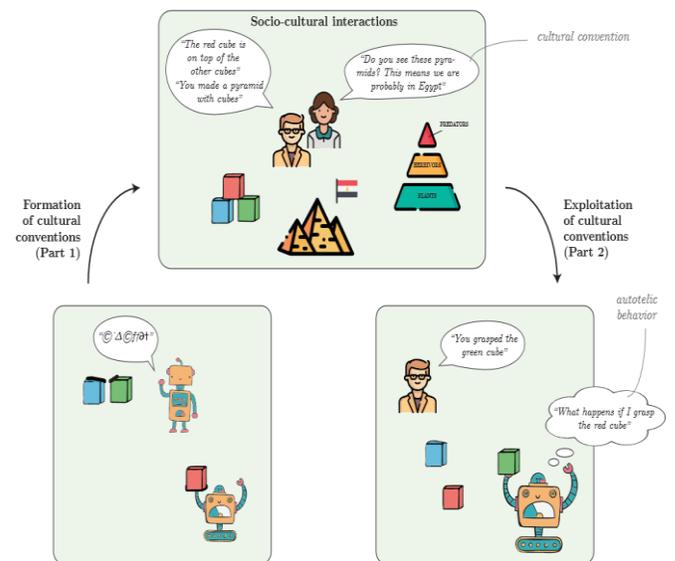

**Fig. 1 Dual organization of the present research. In the first part we take a bottom-up approach and study the self-organization of cultural conventions in artificial agents from social interactions. In the second part, we use a top-down approach to investigate the impact of pre-existing cultural conventions on artificial agents when they interact with social peers**

## 2. Humans are Goal-Directed Social Learners

Humans are an extraordinary inspiration for Artificial Intelligence (AI), as they are the fastest learning system observed. Within just a few years, children learn to crawl, navigate their surroundings, identify and manipulate objects, and even communicate with others. A key element of human development is the concept of goals. As defined by [8], a goal is "a cognitive representation of a future object that the organism is committed to approach or avoid," influencing behaviors significantly. Children's exploratory play is often driven by intrinsically motivated processes that lead them to invent and pursue self-generated goals.

Their exploration is motivated by curiosity and a desire to experience interesting situations, evaluated in terms of optimal incongruity [9]. Moreover, [10] argues that for humans to experience pleasure during learning, they should engage in tasks with the *optimal challenge*, coining the term *autotelic* to describe intrinsically motivated individuals in a state of flow.

## 3. Towards Interactive Social Autonomous Agents

The present research seeks to bridge developmental psychology and modern AI methods to design embodied artificial agents, focusing on "autotelic" and "cultural convention" concepts. Our aim is to create interactive social autotelic agents by immersing them in social contexts and equipping them with mechanisms to construct or exploit cultural conventions. The groundwork for this is built on prior AI paradigms incorporating social elements, such as language-based Reinforcement Learning (RL) [11], instruction-based Imitation Learning (IL) and Multi-Agent RL (MARL) frameworks [12].

Our work contributes by demonstrating that agents can use language as a cognitive tool for goal imagination. This research introduces two experimental contributions, focusing on the self-organization of cultural conventions. First, it explores the role of sensorimotor constraints in forming a graphical language using multi-modal contrastive learning in the context of Language Games [13]. Second, Chapter XVI investigates agent collaboration in the "Architect-Builder Problem," where agents use shared intentionality and pragmatic frames to solve tasks through cultural conventions.

## 4. Background: Standard AI Paradigms

Our contributions bridge standard AI paradigms and developmental psychology to investigate two fundamental research questions (1) the language acquisition problem (self-organization of cultural conventions) and (2) the open-ended skill acquisition problem (self-organization of trajectories). In this chapter, we will first present standard AI problems and their associated families of algorithmic solutions before getting into the specifications of the two problems we investigate.

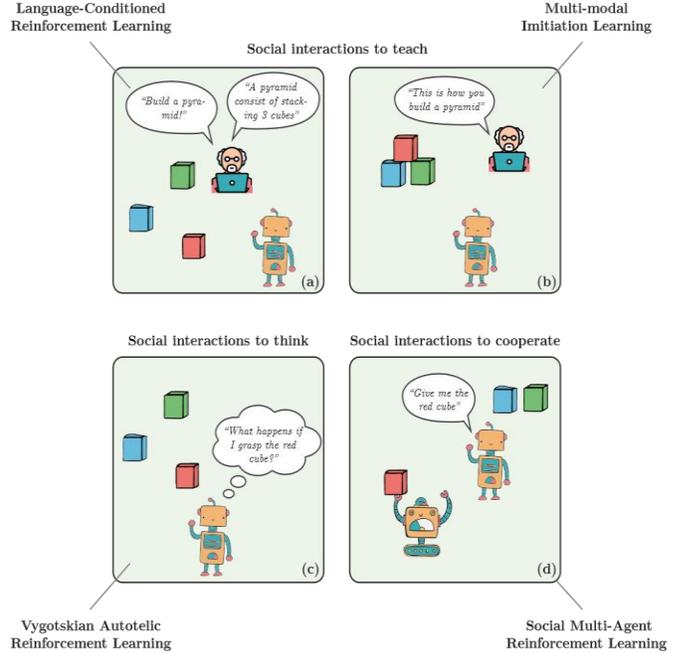

Fig. 2 Social interactions in different AI paradigms. Social interactions and language instructions are used in both RL and IL setting to guide learners. Language can also serve as a cognitive tool to represent goals in autotelic learning. Finally, they can help agents communicate and cooperate in MARL

## 5. Reinforcement Learning
### 5.1. Problem Definition

Reinforcement Learning (RL) involves an agent interacting with an environment to maximize cumulative rewards [1]. Formally, RL is modeled as a Markov Decision Process (MDP), with state space S, action space A, transition function T, initial state distribution $\rho_0$, and reward function $R$. At each time step $t$, the agent selects an action $a_t \in$ A, receives a reward $r_{t+1}$, and observes the next state $s_{t+1} \sim$ T $(s'|s_t, a_t)$. The agent's goal is to learn an optimal policy $\pi^*$ that maximizes expected return:

$$\pi^* = \underset{\pi}{\arg\max}\, \mathbb{E}_{\tau \sim \pi}\left[\sum_{t=0}^{T} \gamma^t R(s_t, a_t)\right]$$

### 5.2. Value Functions

Value functions estimate the expected reward for a state or state-action pair. The state-value function $V_\pi(s)$ and action-value function $Q_\pi(s,a)$ for policy $\pi$ are defined by the Bellman expectation equations:

$V_\pi(s) = \mathbb{E}_{a \sim \pi, s' \sim T}[R(s,a) + \gamma V_\pi(s')],$

$Q_\pi(s,a) = \mathbb{E}_{s' \sim T}[R(s,a) + \gamma Q_\pi(s', \pi(s'))].$

For the optimal policy, the Bellman optimality equations are:

$V^*(s) = \max_a \mathbb{E}_{s' \sim T}[R(s,a) + \gamma V^*(s')],$

$Q^*(s,a) = \mathbb{E}_{s' \sim T}\left[R(s,a) + \gamma \max_{a'} Q^*(s', a')\right].$

## 6. Imitation Learning

Imitation Learning (IL) [14-16] focuses on agents learning in a Markov Decision Process (MDP) without an explicitly defined reward function, instead relying on demonstrations of the task. This approach is particularly beneficial when designing a task-specific reward is challenging. A notable example is self-driving cars, where the complexity of driving makes it impractical to define a reward function, but ample video footage of human drivers is available for training. The typical formalization of the IL problem involves finding a policy that minimizes the divergence between the expert's feature distribution $q_{\pi^*}(\phi)$ and the learner's feature distribution $p_\pi(\phi)$, expressed as:

$$\hat{\pi} = \arg\min_\pi D(q_{\pi^*}(\phi), p_\pi(\phi)), \tag{1}$$

where D is a measure of difference, such as the KullbackLeibler (KL) divergence. A common method to address the IL problem is Behavioral Cloning (BC), which treats imitation learning as a supervised learning task. Given a dataset of trajectories $\mathcal{D} = \{(\tau_i)\}_{i=1}^{N}$ with $\tau = [(s0,a0),...,(sT,aT)]$, the objective is to minimize the cross-entropy loss:

$$L_\pi = - \mathop{\mathbb{E}}_{(s,a) \sim D} [\log \pi(s,a)]. \tag{2}$$

Minimizing this loss is equivalent to minimizing the KL-divergence between the expert's trajectory distribution $P(\tau|\pi^*)$ and the learner's trajectory distribution $P(\tau|\pi)$. However, simple BC may suffer from distributional mismatch, where the learner's policy deviates from the expert's when operating outside the demonstrated state space. To address this, iteratively collecting new expert data is proposed.

Furthermore, BC can only produce a policy that performs at best as well as the expert, which can be limiting if optimal expert trajectories are unattainable, prompting the exploration of Inverse Reinforcement Learning (IRL). IRL seeks to recover an expert's reward function based on observed trajectories, allowing for the development of an optimal policy through RL.

This process, termed Apprenticeship Learning, guarantees that the learned policy is consistent with a learned value function. Various strategies exist for deriving policies in IRL, including feature expectation-based methods, margin-maximization-based methods, and parameterization of the policy by the reward.

Recent approaches have incorporated techniques akin to Generative Adversarial Networks (GAN) to emulate complex behaviors in high-dimensional environments. Although our contributions do not leverage IRL, it remains a significant area of study for surpassing demonstrator performance through advanced ranking methods.

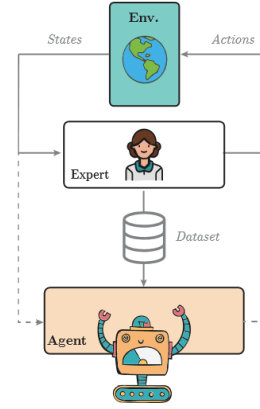

**Fig. 3 Interactions in an IL problem. The agent never interacts with the environment during learning but can interact with it to test its behavior (dashed lines)**

## 7. Multi-Goal Reinforcement Learning

Multi-Goal Reinforcement Learning (MG-RL) extends standard RL by allowing agents to pursue multiple goals framed as constraints over one or more states. Goals can range from specific points to broader subspaces and even language-based goals (e.g., 'find a red object'). A goal-driven agent learns a *goal-conditioned policy* that generates actions based on the current state and goal, formalized as $a_t \sim \pi(\cdot|s_t, z_g)$. In MG-RL, the problem is defined as an MDP with multiple reward functions, $R_G$, where the agent's behavior adapts depending on the goal pursued. Early work on MG-RL led to the development of Universal Value Function Approximators (UVFAs), where a single value function is learned for multiple goals, enabling efficient transfer learning across goals. Techniques like *hindsight learning* further improve sample efficiency by retrospectively using failed trajectories for goal learning.

### 7.1. Typology of Goal Representations

Goal representations vary across tasks. Common approaches include:

- Multiple Targets: One-hot encoded goals with distinct reward functions.
- State Features: Goals defined as target features (e.g., block coordinates, positions) with dense or sparse rewards.
- Dynamic Constraints: Language-based predicates representing constraints to be satisfied.

MG-RL provides a framework for training versatile agents capable of pursuing a diverse set of goals by leveraging shared knowledge across tasks.

## 8. Problem Definition: Formative AI

People possess the ability to surprise, educate, and learn from one another, enabling the transfer and refinement of knowledge across generations. Even without a shared language or prior understanding, such as a parent teaching a child to stack blocks, humans can teach and learn through indirect signals and interactions. Experimental Semiotics studies the forms of communication that emerge when pre-established ones cannot be used, showing that humans can teach and learn without explicit demonstrations or shared protocols. For instance, a CoCo game explored a scenario where an architect guides a builder to construct a structure using arbitrary signals. This raises the question: can artificial agents develop such social conventions?

Inspired by the CoCo game, we propose the Architect Builder Problem (ABP), a framework where learning occurs through social interaction with Markov Decision Processes (MDPs) without direct imitation or reinforcement. The constraints of ABP are: (1) the builder has no prior knowledge of the task, (2) the architect can only communicate via signals, and (3) these signals have no predefined meaning. These challenges make ABP suitable for exploring Human Robot Interaction (HRI), especially in settings like Brain Computer Interfaces (BCIs), where signals and meanings must be learned interactively. To address ABP, we propose Architect-Builder Iterated Guiding (ABIG), an algorithm inspired by shared intent and interaction frames. ABIG allows both agents to learn and refine their communication protocols iteratively.

## 9. Self-Organization Theory

The concept of emerging order stems from chaos theory and describes thermodynamic systems that self-organize from complex interactions. Formalized by cybernetician Ashby, self-organization refers to complex dynamical systems organizing around stable points called 'attractors'. An example is seen in the visual illusion, where perception shifts between two attractors: a young woman or an old woman. Self-organization is evident in nature, such as the formation of sand dunes and snowflakes in physical systems or bee hives and fish schools in biological systems. In technology, this principle enables innovations like adaptive traffic lights.

### 9.1. Self-Organization in Developmental AI

Developmental AI can be framed as adaptive systems where agents and their environment form coupled dynamical systems. This research formalizes two key problems using self-organization:

(1) Language community formation among agents, seen as the self-organization of cultural conventions.

(2) Autonomous skill acquisition as agents self-organize their behavior through internal drivers, leading to developmental trajectories. The autotelic approach integrates social interactions. These problems sit at the intersection of standard and developmental AI, studied further through language formation and the autotelic RL problem.

## 10. Language Game Interactions

Early solutions to the language game involve scoring tables that associate referents with utterances. Agents adjust scores based on communicative success, as illustrated in Figure 4. If predefined categories are unavailable, mechanisms to map visual inputs to object categories have been proposed. The Talking Head experiments adapt language games to dynamic word and meaning inventories. Inspired by the success of Convolutional Neural Networks, extended language games have been extended to image referents, where agents use neural networks for communication. In this framework, a speaker generates an utterance from two images, while a listener selects the target based on the utterance, as shown in Figure 5. Training occurs via Reinforcement Learning (RL) with a reward function reflecting communicative success. Beyond visual referents, researchers analyze the emergence of communication using neural agents. Sequences of symbols can name composite referents, and distinctions between agents' compositional capabilities and the properties of the communication code have been highlighted. Environmental factors influencing compositionality have also been investigated. While guessing interactions serve as an experimental foundation for language formation, human communication encompasses diverse purposes. Thus, AI researchers have examined communication in collaborative tasks using Multi-Agent Reinforcement Learning (MARL). For example, agents have been tasked with car coordination at traffic junctions to avoid collisions.

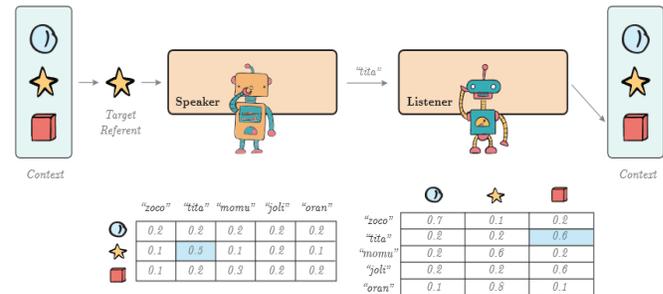

**Fig. 4 Example of agents' tabular internal models, with 3 referents and 5 words**

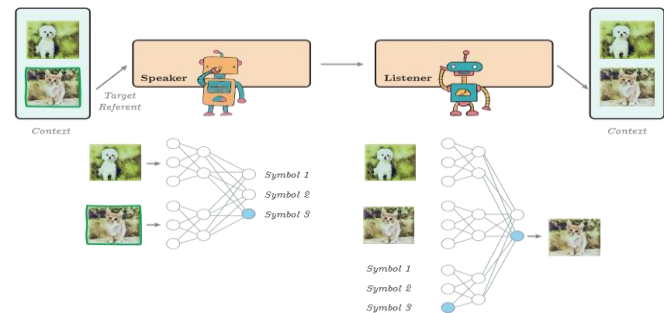

**Fig. 5 Example of agents' neural network internal models, with 2 referents and 3 words**

Two approaches for learning communication in MARL have been introduced: Differentiable Inter-Agent Learning (DIAL) and Reinforced Inter-Agent Learning (RIAL). DIAL enables agents to exchange gradients during centralized training, while RIAL treats messages as actions within an RL framework, allowing agents to communicate without sharing internal states. An attention mechanism has also been added, enhancing agents' communication strategies. This section examined the language formation framework, focusing on linguistic interactions through language games. We demonstrated the scalability of these interactions to neural agents and highlighted MARL's potential for studying communication in collaborative scenarios.

### 10.1. Emergence of Graphical Sensory-Motor Communication

Our first contribution extends the neural communicating agent framework to visual language games via a sensory-motor channel. Unlike prior approaches, which relied on idealized communication channels, we explore whether agents can develop a shared language within a sensory-motor framework. We introduce the Graphical Referential Game (GREG), where a speaker produces graphical utterances to identify visual referents among distractors, illustrated in Figure 7. The utterances are generated using dynamic motor primitives and a sketching library, with referents drawn from the MNIST dataset [17]. Through GREG, we investigate whether agents can self-organise a shared lexicon under sensory-motor constraints, assessing the coherence and compositionality of emerging signals.

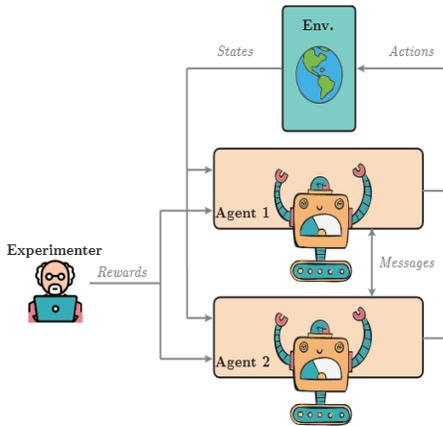

**Fig. 6 Diagram of interactions in MARL emergent communication**

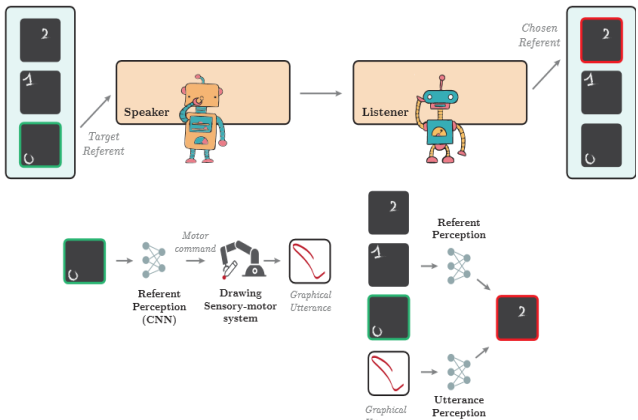

**Fig. 7 The graphical referential game**

### 10.2. The Architect-Builder Problem

Our second contribution introduces the Architect-Builder Problem (ABP), a novel paradigm examining goal-directed communication where the reward function is not accessible to all agents. In this setup, the architect knows the goal and receives rewards but cannot act, while the builder can act but lacks knowledge of the goal. The architect communicates with the builder solely through signals. The ABP addresses gaps in the existing literature by providing a new lens through which to study the emergence of communication in neural agents.

## 11. Self-Organization of Trajectories: The Open-Ended Skill Acquisition Problem

This section examines the self-organization of trajectories in open-ended skill acquisition, where autonomous agents refine skills through environmental interaction. Agents navigate complex action spaces, learning from experiences to improve task performance.

- Learning Objective Distributions: Agents must identify the support of objective distributions—valid goal embeddings. Some methods operate within predefined spaces, while others leverage past representations to shape this distribution using generative models for image-based objectives.
- Goal Selection: After establishing an objective space, agents require strategies for goal selection. Automatic Curriculum Learning (ACL) helps in this regard, coordinating goal sampling for long-term performance improvement. Hierarchical Reinforcement Learning (HRL) sequences goals for lower-level policies, enabling task decomposition.
- Summary: We introduced the autotelic RL framework that fosters intrinsically motivated agents capable of open-ended goal generation.
- Scope: Previous studies using sensory-motor lacked the referential capabilities needed for generating meaningful communication. Our approach, distinguished by stochastic expressions and a decentralized structure, differs significantly in its methods and focus. We investigate factors promoting compositionality in emergent languages, measuring its effectiveness through communicative performance on hidden referents and geometric similarities.

### 11.1. Contributions

This section presents
- The Graphical Referential Game (GREG) for studying sign emergence in a graphical sensory motor framework.
- CURVES: a contrastive multimodal encoder with a generative model for graphical language emergence.
- Performance evaluation of CURVES on unseen structures in various settings.
- Comparative analysis of emerging language structure, focusing on vocabulary stability and compositionality scores via Hausdorff distance.

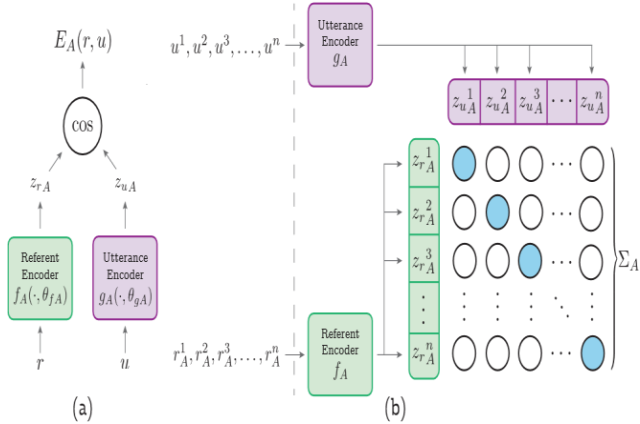

**Fig. 8 (a) Agents' encoder architecture. Referents and expressions map to a shared latent space, where energy is computed as cosine similarity, and (b) Cosine similarity framework update. Agents calculate the energy for all referents and expressions, updating positive samples while minimizing negatives.**

## 12. Graphical Referential Games

We concentrate on referential games, including a speaker ($S$) and an audience ($L$). Each game begins with a set $R$ of $n$ objects (referents) and an objective $r^\star \in R$. The speaker delivers an articulation $u$, and the audience chooses $\hat{r} \in R$. The game succeeds if $\hat{r} = r^\star$.

### 12.1. SetupReferents

Referents are balanced vector highlights (one-hot vectors). For m highlights $F_m$, the referent set is $R_m = \{\sum_{f \in S} f \mid S \subseteq F_m\}$, with subsets of k highlights $\mathcal{R}_m^k = \{\sum_{f \in S} f \mid |S| = k\}$.

Here, $m = 5$.

Highlights from the MNIST dataset are planned by means of $\Phi : R_m \to \tilde{R}_m$. Referents are introduced as $4 \times 4$ frameworks, with specialists seeing alternate points of view: $\tilde{R}_S$ (speaker) and $\tilde{R}_L$ (audience).

#### 12.1.1. Configurations
- One-hot: $r \in R_m$.
- Visual-shared: $r \in \tilde{R}_m$, $\tilde{R}_S = \tilde{R}_L$.
- Visual-unshared: $r \in \tilde{R}_m$, $\tilde{R}_S \neq \tilde{R}_L$.

#### 12.1.2. Sensory-motor Drawing System
Articulations are produced by $M : R^m \to U \subset R^{D \times D}$ utilizing Dynamical Development Natives (DMPs), defined by $c \in R^{20}$. Smooth directions $T = \{v_i\}$ are rendered into $D \times D$ images ($D = 52$) by means of Differentiable Rendering.

### 12.2. Objectives We Address
- Can the agent solve the game and generalize to compositional referents?
- Are the articulations interpretable and consistent?
- Do articulations display compositional principles?

Performance: Specialists' training referents:
$R_{train} = \mathcal{R}_5^1$; testing: $\mathcal{R}_{test} = \mathcal{R}_5^2$.

Interpretability: Hausdorff distance $d_H(T_1, T_2)$ assesses similarity:
$d_H(T_1, T_2) = \max\{\sup_{v \in T_1} d(v, T_2), \sup_{v' \in T_2} d(T_1, v')\}$.

Measurements:
- Expert Knowledge: Similarity of articulations for a referent.
- Viewpoint Clarity: Similarity of articulations across viewpoints.
- Referent Clarity: Similarity of articulations across referents.

Compositionality: Compositional score $\rho$ measures articulation comparability for compositional referents $u(r_{ij})$ comparative with $u(r_i)$ and $u(r_j)$.

## 13. CURVES: Contrastive Articulation Referent Agreeable Scoring

It is an energy-based approach with two parts:
- Contrastive learning of an energy scene $E(r,u)$ by means of cosine comparability.
- Articulation age amplifying energy for $r_S^\star$.

### 13.1. Agents and Collaboration

Specialists $A \in \{A_1, A_2\}$ utilize unmistakable CNN encoders $f_A$ (referents) and $g_A$ (articulations), planning to a common $d$-layered space: $z_{rA} = f_A(r)$, $z_{uA} = g_A(u)$. Energy scene: $E_A(r,u) = \cos(f_A(r), g_A(u))$.

The result is $o = 1_{[\hat{r}=r^\star]} - b$, where $b$ is the benchmark achievement rate.

Contrastive Learning: Specialists register closeness grids $\Sigma_A$:
$(\Sigma_A)_{i,j} = E_A(r_A^i, u^j)$,
with the goal:
$$J_A(\Sigma_A, I) = \frac{CE((\Sigma_A)_{i,1:n}, e_i) + CE((\Sigma_A)_{1:n,i}, e_i)}{2},$$

Where $e_i$ is a one-hot vector. Speaker and audience misfortunes:
$$\min_{\theta_{f_S}, \theta_{g_S}} \sum_{i=1}^{n} o_i J_S(\Sigma_S, I), \quad \min_{\theta_{f_L}, \theta_{g_L}} \sum_{i=1}^{n} J_L(\Sigma_L, I)$$

### 13.2. Expression Generation: Two Strategies
- Distinct: Boost cosine comparability for $r_S^\star$:
  $c^\star = \underset{c \in R^p}{\arg\max}\, E(r_S^\star, M(c))$.

- Discriminative: Limit cross-entropy for $r_S^\star$:

$c^\star = \underset{c \in \mathbb{R}^p}{\operatorname{argmin}}\ CE(\sigma_S, e_r S\star)$,

Where $\sigma_S$ contains comparability scores for $\tilde{R}_S$.

## 14. Experiments
### 14.1. Communicative Performance
Specialists made close amazing preparation progress rates (SR) across every game setting (one-hot, visual-shared, and visual-unshared).

#### 14.1.1. Generalization to Compositional Referents
Table 1 shows speculation execution on compositional referents ($r \in \mathcal{R}_5^2$) inside thorough settings (|R| = 10). Standard correlations incorporate a random technique ($SR_{random} = 0.1$) and a single-feature procedure ($SR_{1\text{-feature}} = 0.25$).

Specialists performed well across referent sorts, with one-hot settings yielding the most elevated SR, while execution in visual settings declined because of added intricacy according to point-of-view shifts. Curiously, enlightening and discriminative expressions made comparable progress rates, demonstrating that limiting uncertainty in articulations doesn't essentially support speculation execution.

### 14.2. Emergent Language Structure
#### 14.2.1. Coherence
Specialists exhibited expanding between specialist, between the point of view and between referent lucidness during preparation. Higher intelligence scores compared to merging correspondence conventions. For visual referents, particular signs arose, showing specialists' capacity to really separate referents.

#### 14.2.2. Compositionality
While specialists accomplished high SR for one-hot compositional referents, the produced articulations needed clear compositional designs. Examinations utilizing mathematical mappings, like distance measurements, showed a restricted vicinity between signals for compositional and individual referents, as portrayed in Figure 10.

#### 14.2.3. Conclusion
Regardless of the shortfall of unequivocal compositional designs in created articulations, inward portrayals recommend specialists use compositional procedures. Compelled test arrangements are fundamental for additional investigation of rising language properties.

**Table 1. Generalization success rates for compositional referents, evaluated across different game settings**

| Setting | Descriptive SR | Discriminative SR |
|---|---|---|
| One-hot | .99± .01 | .99± .01 |
| Visual-shared | .57± .03 | .56± .03 |
| Visual-unshared | .39± .02 | .40± .02 |

**Table 2. Training and test success rates without perspective variability**

| Setting | Training SR | Test SR |
|---|---|---|
| One-hot | .99± .01 | .96± .02 |
| Visual-shared | .99± .01 | .55± .03 |
| Visual-unshared | .99± .01 | .41± .02 |

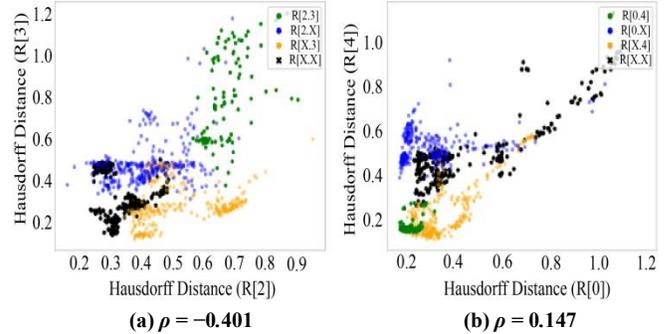

(a) $\rho = -0.401$   (b) $\rho = 0.147$
**Fig. 9 Topographic map models for a solitary seed in onehot referents setting**

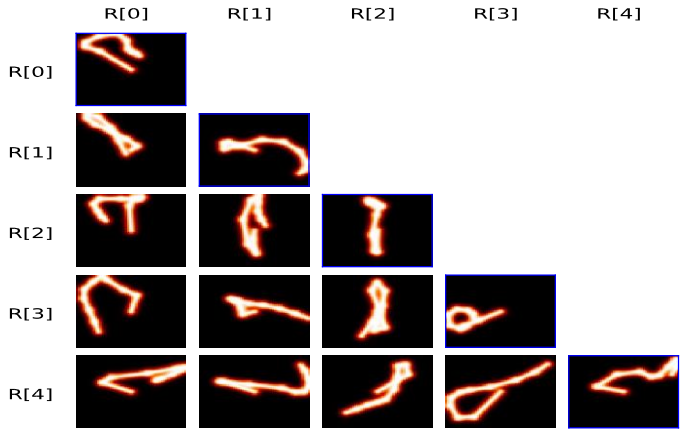

**Fig. 10 Matrix of arrangements**

Every expression names a compositional referent and is shaded in blue in the event that it contains highlight $i$ ($R[i,X]$), orange in the event that it contains include $j$ ($R[X,j]$), green assuming it contains both ($R[i,j]$), and dark assuming it contains none ($R[X,X]$). (a) Relating to the most terrible geographical score $\rho = -0.401$ (mix of component $i = 2$ and $j = 3$) (b) Comparing to the best geological score $\rho = 0.147$ (mix of element $i = 0$ and $j = 4$). Blue outlines address expressions created for a viewpoint in $\mathcal{R}_5^1$, and other expressions signify the comparing syntheses in $\mathcal{R}_5^2$

## 15. Experimental Design and Evaluation Metrics
To evaluate the efficacy of autotelic Reinforcement Learning (RL), a structured experimental setup was devised. The primary objective was to assess the agents' ability to autonomously generate and master diverse goals in dynamic environments. The experimental design encompassed the following key aspects:

Environment Configuration: Agents were trained in simulated environments characterized by varying complexity levels. These environments included tasks requiring object manipulation, maze navigation, and abstract problem-solving, ensuring a comprehensive evaluation of agent capabilities.

Evaluation Metrics:
- Exploration: Quantified using diversity measures that assess the range of goals generated and achieved by the agent.
- Generalization: Evaluated by testing agents on unseen goals and scenarios, measuring success rates compared to training tasks.
- Robustness: Assessed through perturbation tests, where environmental variables were systematically altered to determine the agent's adaptability.

Implementation Details: The experimental framework incorporated various RL algorithms as baselines, enabling a comparative analysis of the autotelic framework against traditional approaches. Hyperparameter settings, training iterations, and computational resources were standardized across all experiments for consistency. This detailed experimental framework ensures that the proposed metrics capture the multifaceted performance of autotelic RL agents, providing a robust foundation for assessing their capabilities.

## 16. Discussion

In this part we formalized GREG: another biological referential game where two specialists should convey by means of a persistent tangible engine framework emulating a mechanical arm drawing outlines. To handle GREG, we propose CURVES: a contrastive portrayal of learning calculation enlivened by early language game contrastive execution that scales to high-layered signals. CURVES permits a gathering of two specialists two combine on a common graphical language in settings where referents are one-hot vectors or pictures of MNIST digits. The portrayals that specialists learn to empower them to convey compositional referents never experienced during preparation. Assuming the Haussdorf distance shows that rising signs are intelligent, it doesn't catch compositionality among them. Future work might use our natural arrangement and algorithmic answer to try different things with and test various theories that impact structures in self-coordinating using frameworks. An examination of the effect of the tangible engine imperatives on the geography of graphical signs could, for example, give a significant understanding of the natural elements working with the development of a compositional graphical language. Motivated by work on the social development of language, our arrangement can likewise act as a premise to explore and imagine the effect of different factors like populace dynamics or the mental capacities of specialists (with fluctuating memory or perceptual frameworks). At long last, CURVES is rationalist to the methodology used to address expressions. All things considered, it could handle other tactile engine frameworks. The focal component of CURVES lies in the contrastive learning of expression referent affiliations. In our execution, we improve expressions by boosting this energy through angle climb. Similar to the Clasp that opened numerous roads for multi-modular age, we could connect more intricate generative techniques like dispersion models.

## 17. Learning to Guide and to be Directed in the Draftsman Developer Problem

In this section, we explore the emergence of goal-directed communication between artificial agents in a novel setting, contrasting with the classical referential game where agents share the reward function. Specifically, we study collaboration between a builder - who performs actions but lacks access to rewards - and an architect - who guides the builder towards the task's goal. This scenario requires agents to learn a task while simultaneously developing a communication protocol without predefined meanings. Drawing inspiration from Experimental Semiotics, we introduce the Architect-Builder Problem (ABP), where the architect knows the goal but can only send messages, while the builder acts without knowing the task, relying on the architect's guidance. We propose Architect Builder Iterated Guiding (ABIG), where the architect uses a learned model of the builder to guide it, and the builder employs self-imitation learning to reinforce its behavior. ABIG organizes interactions into structured frames, enabling agents to develop a reusable communication protocol, tested in a 2D environment with tasks like grasping cubes and building shapes, demonstrating effective generalization to unseen tasks.

### 17.1. Sociocultural Dynamics in Autotelic RL

Socio-cultural dynamics underpinning autotelic RL draw inspiration from human learning, where interactions with peers and the environment significantly influence skill acquisition. This study explores the integration of these dynamics into artificial agents, leveraging frameworks from developmental psychology and cognitive science.

- Theoretical Foundations: Sociocultural theories, such as Vygotsky's concept of the Zone of Proximal Development (ZPD), highlight how collaboration and shared knowledge accelerate learning
- By embedding similar principles into RL agents, the study aims to foster goal imagination and refinement through simulated peer interactions.
- Agent Interactions: Agents were immersed in environments where they could interact with other agents or simulated humans, exchanging information and adopting sociocultural conventions. For example, agents used language-like constructs to negotiate tasks and shared strategies for goal achievement.
- Implications for Learning: Sociocultural interactions enhanced the agents' ability to:
  - Generate novel goals inspired by peer actions.

- Learn more efficiently by observing and imitating successful strategies.
- Adapt to changing environmental dynamics by internalizing shared conventions.

This subsection provides a deeper understanding of how sociocultural factors can be operationalized in autotelic RL frameworks, advancing the design of autonomous agents capable of emulating human-like adaptability and creativity.

## 18. Motivations

People possess the ability to surprise, educate, and learn from one another, enabling the transfer and refinement of knowledge across generations. Even without a shared language or prior understanding, such as a parent teaching a child to stack blocks, humans can teach and learn through indirect signals and interactions. Experimental Semiotics studies the forms of communication that emerge when pre-established ones cannot be used, showing that humans can teach and learn without explicit demonstrations or shared protocols. For instance, explored a CoCo game where an architect guides a builder to construct a structure using arbitrary signals. This raises the question: can artificial agents develop such social conventions?

Inspired by the CoCo game, we propose the Architect Builder Problem (ABP), a framework where learning occurs through social interaction with Markov Decision Processes (MDPs) without direct imitation or reinforcement. The constraints of ABP are:
(1) The builder has no prior knowledge of the task.
(2) The architect can only communicate via signals.
(3) These signals have no predefined meaning.

These challenges make ABP suitable for exploring Human Robot Interaction (HRI), especially in settings like Brain Computer Interfaces (BCIs), where signals and meanings must be learned interactively. To address ABP, we propose Architect-Builder Iterated Guiding (ABIG), an algorithm inspired by shared intent and interaction frames. ABIG allows both agents to iteratively learn and refine their communication protocols.

## 19. Designer Developer Problem
### 19.1. Planner Developer Issue

We consider a multispecialist arrangement made out of two specialists: a planner and a developer. The two specialists notice the climate state $s$, yet only the planner knows the objective within reach. The engineer can't make moves in the climate yet gets the natural prize $r$, though the manufacturer gets no award and has, in this manner, no information about the job that needs to be done. In this uneven arrangement, the modeler can collaborate with the developer through a correspondence signal $m$ examined from its strategy $\pi_A(m|s)$. These messages, which have no deduced implications, are gotten by the developer, which acts as per its approach $\pi_B(a|s,m)$. This makes the climate change to another state $s'$ tested from $P_E(s'|s,a)$, and the designer gets reward $r'$. Messages are sent at each time step. The CoCo game that propelled ABP is outlined in Figure 11(a), while the general designer developer climate connection chart is given in Figure 11(b). The distinctions between the ABP setting and the MARL and IRL settings are represented. The engineer and the developer ought to team up to assemble the development focus while situated in various rooms. The engineering has an image of the objective while the manufacturer approaches the blocks. The planner screens the manufacturer's work area by means of a camera (video transfer) and can speak with the developer just using 10 images (button occasions). (b) Interaction chart between the specialists and the climate in our proposed ABP. The draftsman conveys messages ($m$) to the manufacturer. Just the developer can act ($a$) in the climate. The developer conditions its activity on the message sent by the manufacturer ($\pi_B(a|s,m)$). The manufacturer never sees any compensation for the climate. A schematic perspective on the identical ABP issue.

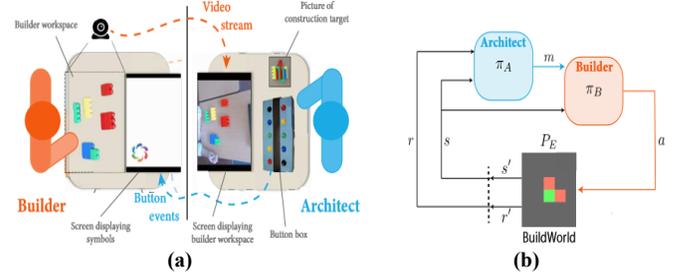

**Fig. 11(a) Schematic perspective on the CoCo Game (the motivation for ABP), (b) Designer developer climate connection chart**

BuildWorld: We direct our trials in BuildWorld. BuildWorld is a 2D development framework universe of size (w × h). Toward the start of an episode, the specialist and $N_b$ blocks are produced at various arbitrary areas. The specialist can explore this world and handle blocks by enacting its gripper while on a block. The activity space A is discrete and incorporates a "sit idle" activity (|A| = 6). At each time step, the specialist notices its situation in the framework, its gripper state, as well as the place of the relative multitude of blocks and on the off chance that they have gotten a handle on (|S| = 3 + 3$N_b$). Tasks. BuildWorld contains 4 different preparation errands: 'Handle': The specialist should get a handle on any of the blocks; 'Spot': The specialist should put any block at a predefined area in the lattice; 'H-Line': The specialist should put every one of the blocks in a flat line design; 'V-Line':The specialist should put every one of the blocks in an upward line design. BuildWorld likewise has a harder fifth testing task, '6-blocksshapes', that comprises additional mind-boggling designs, and that is utilized to challenge a calculation's exchange capacities. For all assignments, rewards are inadequate and possibly given when the undertaking is finished. This climate typifies the intuitive learning challenge of ABP while eliminating the requirement for complex insight or movement. In the RL setting, where a similar specialist acts and gets rewards, the specialist that demonstrates in the climate is likewise the one that gets the ground-truth reward. This climate wouldn't be exceptionally noteworthy. In any case, it still needs to be demonstrated the way that the undertakings can be settled in the testing learning setting of ABP (with a prize less developer and an activity-less modeler).

### 19.2. Communication
The engineer directs the developer by sending messages $m$ which are one-hot vectors of size $|V|$ going from 2 to 72.

### 19.3. Additional Suppositions
To zero in on the planner developer cooperations and the learning of a common correspondence convention, the draftsman approaches $P_E(s'|s,a)$ and to the prize capability $r(s,a)$ of the current objective. This is expected to be that, assuming the planner were to act in the climate rather than the manufacturer, it would have the option to rapidly sort out some way to address the errand. This supposition is viable with the CoCo game trial where people members, and specifically the engineers, are known to have such world models.

## 20. ABIG: Draftsman Manufacturer Iterated Guiding
### 20.1. Analytical Description
Agents-MDPs. In the Planner-Builder problem, agents operate in distinct but coupled MDPs based on their perspectives. For the planner, messages act as actions that influence both the next state and reward. The planner knows the environment's transition function $P_E(s'|s,a)$ and the reward function $r(s,a)$, which are independent of messages. This allows the planner to predict how its messages will affect the builder's actions, guiding the reward and subsequent states. The builder's state, however, includes both the environment state and the message, making state transitions more complex as message dynamics must be captured. Despite this, the builder can leverage its knowledge of the planner's message choices, which are based on the current environment state.

Under shared planning assumptions, planner-builder interactions, where the planner optimizes its policy $\pi_A^*$ to maximize $G_A$, will also maximize $G_B$. The builder can interpret these interactions as demonstrations that maximize its unknown reward function $\tilde{r}$. By performing self-imitation Learning on these interaction trajectories $\tau$, the builder can reinforce its behavior towards the planner's guidance, making the builder's MDP non-stationary. To address this, agents rely on interaction frames where one agent's policy is fixed while the other learns, restoring stationarity. The planner's MDP is defined as $M_A = \langle S, V, P_A, r_A, \gamma \rangle$, and the builder's MDP as $M_B = \langle S \times V, A, P_B, \emptyset, \gamma \rangle$, where $\pi_A$, $\pi_B$, $P_A$, $P_B$, and reward functions are defined as above.

### 20.2. Practical Algorithm
ABIG iteratively structures interactions between an architect builder pair into interaction frames. Each cycle starts with a modeling frame, where the architect learns a model of the builder's behavior. This is followed by the guiding frame, where the architect uses the learned model to produce messages that guide the builder. The builder stores these interactions to refine its policy $\pi_B$ through self-imitation. The guiding frame involves the architect using Monte Carlo Tree Search (MCTS) to generate optimal messages based on simulated builder reactions. The builder then updates its policy with Behavioral Cloning (BC) on stored interactions $D_B$.

ABIG is general and can handle various tasks without limiting the type of communication protocol that emerges. We also investigate two control settings: ABIG -no-intent, where the builder interacts with an architect sending random messages during training, and random, where the builder takes random actions. These control settings help measure the impact of self-imitation during guiding versus non-guiding interactions, with random actions providing a performance lower bound. All models use two-layer 126-unit ReLu networks, and the architect's MCTS employs UCT heuristics for further details on training and hyperparameters. BC optimizes the cross-entropy loss using the Adam optimizer.

Specialists iteratively cooperate through the displaying and directing edges. In each edge, one specialist gathers information and works on its arrangement while the other specialist's way of behaving is fixed. Modeling Casing: The modeler records an informational index of collaboration.

## 21. Conclusion and Future Work
This work formalizes the Architect-Builder Problem (ABP) as an intelligent setting where agents learn to collaborate without explicit support, examples, or a shared language. To address the ABP, we propose the Architect-Builder Interaction Mechanism (ABIM), a method that enables agents to both guide and be guided. ABIM relies on two high-level priors for communication development: shared intentionality and interaction frames. The flexible design of ABP allows us to enforce these priors during learning formally. Through ablation studies, we highlight the critical role of shared intentionality, achieved via self-imitation of guiding actions. When applied in interaction frames, this mechanism enables agents to develop a communication protocol that allows them to solve all tasks in BuildWorld. Remarkably, we find that communication protocols learned on simpler tasks can be extended to tackle harder, unseen challenges. Despite its effectiveness, our approach has some limitations that present opportunities for future research.

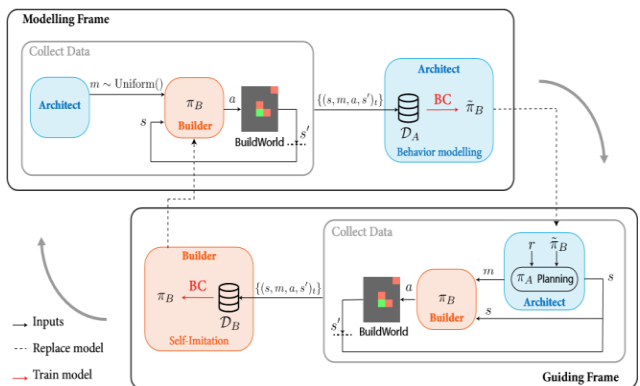

**Fig. 12 Architect-developer iterated guiding**

First, ABIM trains agents in a fixed interaction framework, which involves several structured episodes, making it data-inefficient. A potential direction for improvement is to relax this stationarity assumption and allow agents to learn from dynamic, non-stationary data buffers containing past behaviors. Second, the builder remains dependent on the architect's guidance even during convergence. A Vygotskian approach could allow the builder to internalize the architect's guidance, becoming more autonomous. For instance, the builder could learn a model of the architect's messaging strategy after communication protocols stabilize. Further research could explore memory mechanisms to facilitate feedback loops, experiment with low-frequency feedback, or investigate compositional message structures.

Ultimately, ABP provides a testbed for studying the fundamental mechanisms of emergent communication and the impact of high-level communication priors from experimental semiotics.

**Algorithm 1:** Architect-Manufacturer Iterated Directing (ABIG)
**Require :**
haphazardly introduced manufacturer strategy $\pi_B$, reward capability $r$, change capability $P_E$, BC calculation, MCTS calculation
for $i$ in range($N_{iterations}$) do
**Modelling Frame:**
for $e$ in range($N_{collect}/2$) do Engineer populates $D_A$ utilizing $m \sim$ Uniform() and noticing $a \sim \pi_B(\cdot|s,m)$
end for
Engineer learns $\tilde{\pi}_B(a|s,m)$ on $D_A$ with BC Engineer sets $\pi_A(m|s) \triangleq$ MCTS($r, \tilde{\pi}_B, P_E$)
Engineer flushes $D_A$
**Guiding Frame:**
for $e$ in range($N_{collect}/2$) do manufacturer populates $D_B$ utilizing $\pi_B$ while directed by designer, for example $m \sim \pi_A(\cdot|s)$
end for
Manufacturer learns $\pi_B(a|s,m)$ on $D_B$ with BC
Manufacturer flushes $D_B$ end for
Designer runs one final Demonstrating frame
Result: $\pi_A, \pi_B$